\documentclass[conference]{IEEEtran}
\IEEEoverridecommandlockouts
\usepackage{cite}
\usepackage{amsmath,amssymb,amsfonts}
\usepackage{graphicx}
\usepackage{textcomp}
\usepackage{algorithm}
\usepackage{algpseudocode}
\usepackage{multicol}
\usepackage{booktabs}
\usepackage{multirow}
\usepackage{amsmath}
\usepackage{bm}
\usepackage{amssymb}
\usepackage{cite}
\usepackage{tabularx}

\usepackage{graphicx}
\usepackage{newfloat}
\usepackage{listings}
\usepackage{xcolor}
\def\BibTeX{{\rm B\kern-.05em{\sc i\kern-.025em b}\kern-.08em
    T\kern-.1667em\lower.7ex\hbox{E}\kern-.125emX}}
\begin{document}

\title{Knowledge Rumination for Client Utility Evaluation in Heterogeneous Federated Learning\\

\thanks{* Corresponding authors. This work is supported by the National Key Research and Development Program of China (2022YFB3105405, 2021YFC3300502) and the Provincial Key Research and Development Program of Anhui (202423110050033).} 
}

\author{\IEEEauthorblockN{Xiaorui Jiang}
\IEEEauthorblockA{CCCD Key Lab of Ministry of  Culture\\
and Tourism, University of Science
and\\ Technology of China, Hefei, China \\
xrjiang@mail.ustc.edu.cn}
\and
\IEEEauthorblockN{Yu Gao}
\IEEEauthorblockA{CCCD Key Lab of Ministry of  Culture\\
and Tourism, University of Science
and\\ Technology of China, Hefei, China \\
yugao@mail.ustc.edu.cn}
\and
\IEEEauthorblockN{Hengwei Xu}
\IEEEauthorblockA{CCCD Key Lab of Ministry of  Culture\\
and Tourism, University of Science
and\\ Technology of China, Hefei, China \\
xuhw@mail.ustc.edu.cn}
\and
\IEEEauthorblockN{Qi Zhang}
\IEEEauthorblockA{Department of Electrical and
Computer\\ Engineering, Aarhus University, \\ Aarhus, Denmark\\
qz@ece.au.dk}
\and
\IEEEauthorblockN{Yong Liao*}
\IEEEauthorblockA{CCCD Key Lab of Ministry of  Culture\\
and Tourism, University of Science
and\\ Technology of China, Hefei, China \\
yliao@ustc.edu.cn}
\and
\IEEEauthorblockN{Peng Yuan Zhou*}
\IEEEauthorblockA{Department of Electrical and
Computer\\ Engineering, Aarhus University, \\ Aarhus, Denmark\\
pengyuan.zhou@ece.au.dk}
}

\maketitle

\begin{abstract}
Federated Learning (FL) allows several clients to cooperatively train machine learning models without disclosing the raw data. In practical applications, asynchronous FL (AFL) can address the straggler effect compared to synchronous FL. However, Non-IID data and stale models pose significant challenges to AFL, as they can diminish the practicality of the global model and even lead to training failures. In this work, we propose a novel AFL framework called Federated Historical Learning (FedHist), which effectively addresses the challenges posed by both Non-IID data and gradient staleness based on the concept of knowledge rumination. FedHist enhances the stability of local gradients by performing weighted fusion with historical global gradients cached on the server. Relying on hindsight, it assigns aggregation weights to each participant in a multi-dimensional manner during each communication round. To further enhance the efficiency and stability of the training process, we introduce an intelligent $\ell_2$-norm amplification scheme, which dynamically regulates the learning progress based on the $\ell_2$-norms of the submitted gradients. Extensive experiments indicate FedHist outperforms state-of-the-art methods in terms of convergence performance and test accuracy.
\end{abstract}

\begin{IEEEkeywords}
federated learning, stale gradients, non-iid data
\end{IEEEkeywords}

\section{Introduction}
\label{sec:intro}

As a distributed machine learning paradigm, Federated learning (FL) \cite{mcmahan2017communication} has garnered widespread attention in recent years due to its ability to protect the privacy of end-user data. In a classical FL workflow, each client uses its private data to train a local model and then uploads the gradient to a central server. The central server aggregates the collected local gradients into a global gradient to update the global model, which is then sent back to the clients for subsequent training rounds. The entire process does not leak the raw data. As such, FL effectively safeguards data privacy and provides unique advantages in various domains including healthcare \cite{brisimi2018federated}, autonomous driving \cite{liang2022federated}, and mobile keyboard prediction \cite{hard2018federated}. A well-known FL aggregation algorithm is federated averaging (FedAvg) \cite{mcmahan2017communication}, where the server aggregates updates from a selected group of local clients and computes the global model by applying a weighted averaging strategy. Due to the system and statistical heterogeneity among devices, this approach often encounters the straggler effect \cite{cipar2013solving}, which severely hampers the training progress. In response, \cite{xie2019asynchronous} introduced asynchronous FL (AFL), where the server no longer waits for all clients but can promptly update the global model upon receiving a single local gradient. However, AFL may lead to global model bias towards fast clients, and the frequent data transmissions may result in server crashes \cite{shi2020device}. Some studies have tried to seek a compromise between synchronous and asynchronous approaches \cite{wu2020safa,zhou2022towards}, where K-async FL \cite{zhou2022towards} has demonstrated promising results, particularly in highly heterogeneous environments. In K-async FL, the server initiates aggregation as soon as it receives the first $K$ local gradients, while the remaining clients continue training.

\begin{figure}[t]
\centering
\includegraphics[width=1.0\linewidth]{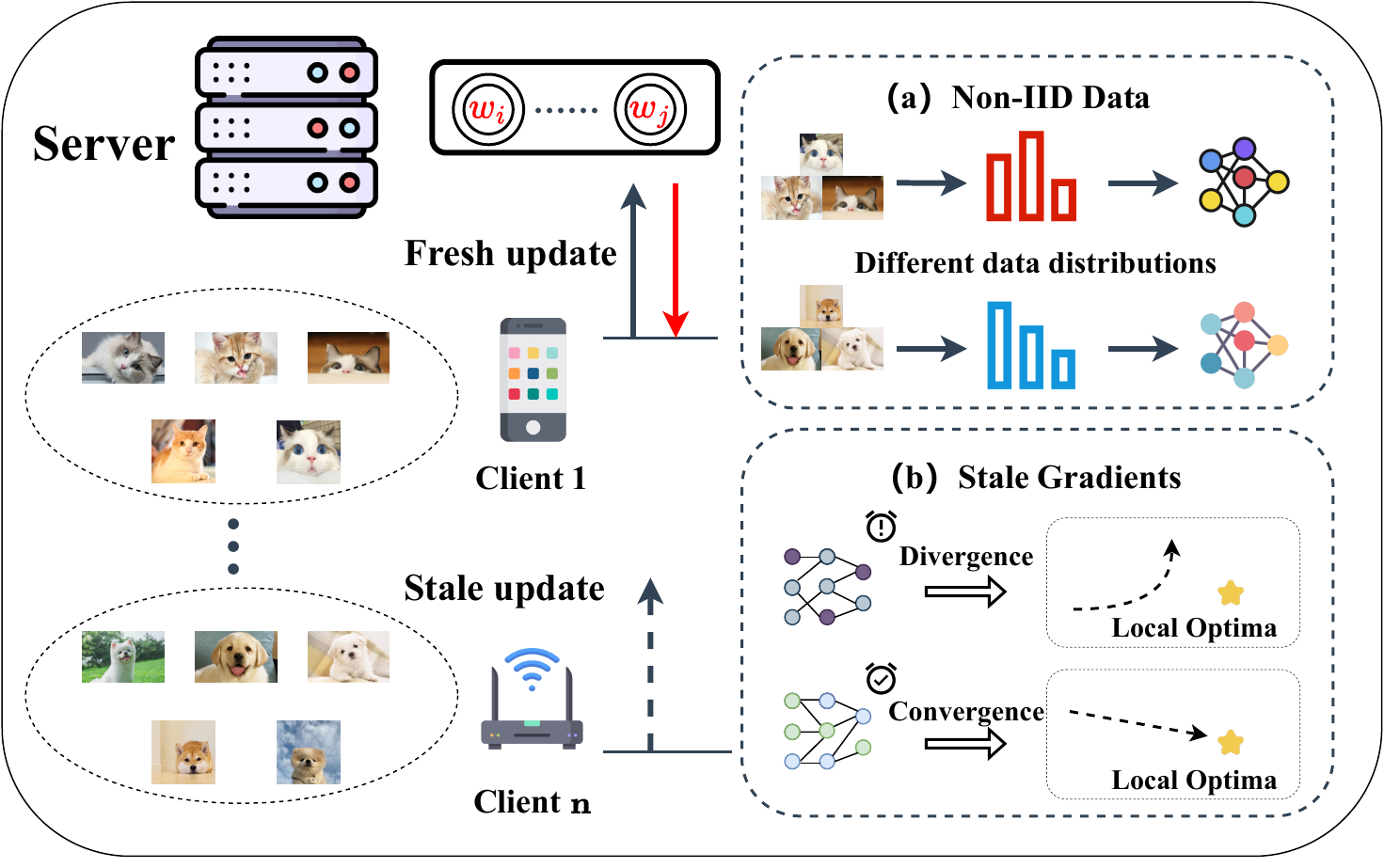}
\caption{Problem illustration of K-async FL. (a) Non-IID data introduces biases among local gradients. This heterogeneity affects the convergence of the global model.
(b) Stale gradients are often detrimental to the global model, causing update directions to deviate from local optima.} 
\label{problem}
\end{figure}

Despite the favorable performance of K-async FL, it is still constrained by two challenges, as shown in Figure \ref{problem}. On the one hand, the presence of non-independent and identically distributed (Non-IID) data poses a significant challenge in training a well-performing global model \cite{zhao2018federated}. On the other hand, the existence of stale local gradients often affects the utility of the global gradient, even leading to divergence \cite{zhang2023timelyfl}. Currently, it is common practice to adopt strategies such as variance reduction \cite{karimireddy2020scaffold,li2021model}, regularization \cite{li2020federated}, and momentum \cite{li2019gradient,xu2022coordinating} to accommodate heterogeneous scenarios. As for the latter, the mainstream approach involves assigning different weights to local gradients based on their staleness \cite{chen2019communication,damaskinos2022fleet} to alleviate the weight divergence.
%
While some algorithms can address both aforementioned issues simultaneously, they either rely on a sole weighted criterion \cite{zhou2022towards} leading to a biased allocation of weights, or necessitate specific prior knowledge that is not readily accessible such as the grouping of clients based on varying computational capacities \cite{li2022fedhisyn}, leading to challenges in practical application. 

\begin{figure*}[!t]
\centering
\includegraphics[width=1.0\linewidth]{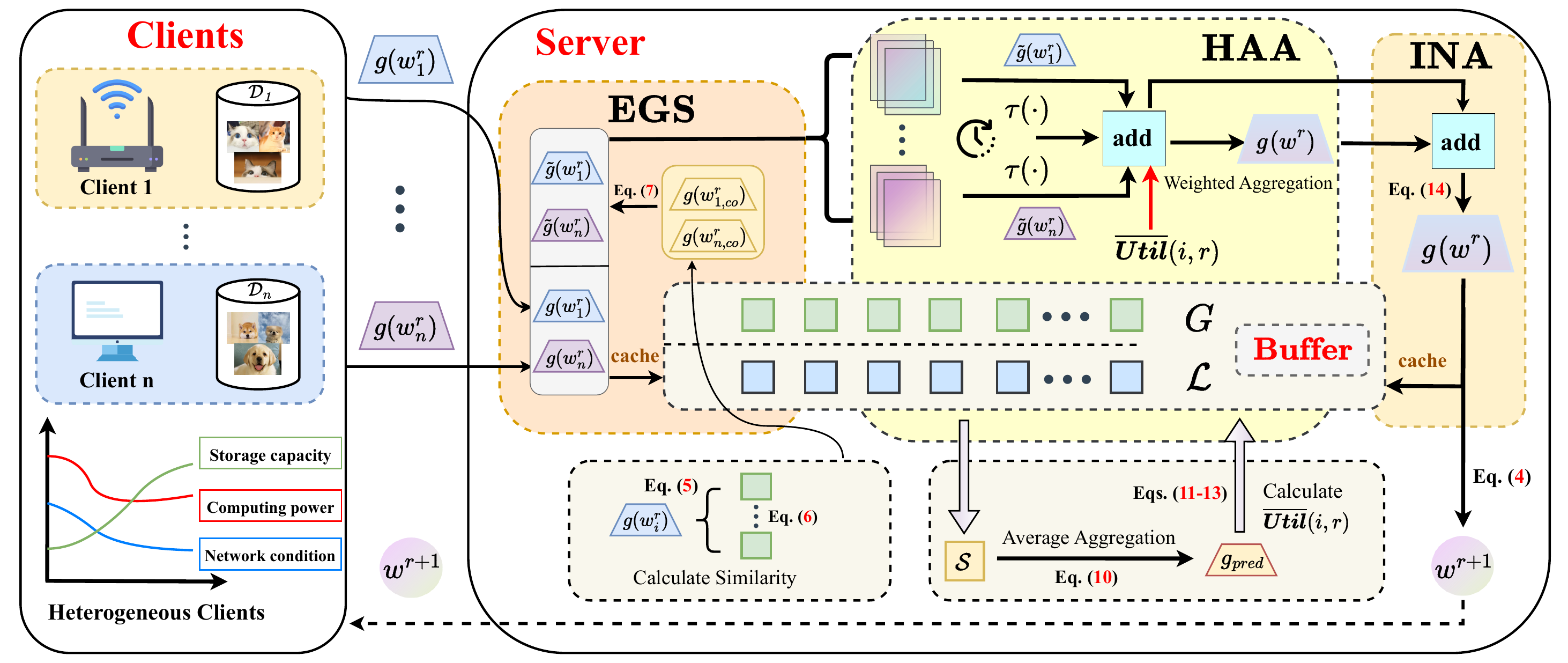}
\caption{The overall framework of FedHist. It comprises three components: EGS, HAA, and INA.}
\label{framework}
\end{figure*}

To address these two challenges, drawing inspiration from the concept of \underline{Knowledge Rumination}, we propose FedHist (the framework is illustrated in Figure \ref{framework}), a novel K-async FL framework that is capable of handling Non-IID data and stale gradients simultaneously. \textbf{Knowledge Rumination} refers to a behavior, akin to certain animals regurgitating food from the stomach back to the oral cavity for rechewing, aimed at fully absorbing nutrients. Specifically, we leverage a historical gradient buffer at the server-side to facilitate knowledge sharing across clients, enhancing the adaptability to Non-IID data. And we use the historical updates of clients to evaluate their utilities, as a factor for determining their weights during the near future aggregation stage. It is noteworthy that the primary distinction of FedHist from previous works lies in our innovative use of historical updates to evaluate client utility, resolving asynchronous issues from a synchronous perspective. Additionally, we investigate factors that influence convergence and propose a strategy for amplifying the $\ell_2$-norms of aggregated gradients adaptively to accelerate the convergence of K-async FL. More importantly, FedHist does not require any prior knowledge about the clients, such as computational capabilities, and maintains the same level of privacy as FedAvg. In summary, FedHist effectively addresses the challenges posed by Non-IID data and stale gradients in AFL.
Our main contributions can be summarized as follows:

\begin{itemize}
    \item We propose FedHist, a novel solution tailored for Non-IID data and stale gradients in AFL, deftly accommodating both of these challenges.
    \item We introduce a gradient buffer to achieve history-aware aggregation on the basis of improving the stability of local gradients and propose a $\ell_2$-norm amplification strategy. To the best of our knowledge, we are the first to introduce the concept of knowledge rumination for client utility evaluation, thereby enhancing the effectiveness.
    \item  Extensive experiments on public datasets demonstrate that FedHist outperforms state-of-the-art methods and effectively promotes fairness in the training process.
\end{itemize}

\section{Methodology}

\subsection{Problem Statement}
Suppose there are $N$ clients, denoted as $c_1,c_2,\ldots,c_N$, and each client $c_i$ trains a model using its local dataset $\mathcal{D}_i = \{ \xi_{i,1},\xi_{i,2},\cdots,\xi_{i,\lvert{\mathcal{D}_i} \rvert } \}$, where $\xi_{i,j}$ represents the $j$-th data point $\{x_j,y_j\}$ on client $i$. Our objective is to collaboratively train a global model $w$ using the entire dataset $\mathcal{D}=\bigcup_{i \in N} \mathcal{D}_i$ without compromising the privacy of the raw data. For client $c_i$, its empirical loss is defined as: 

\begin{equation} \label{eq1}
F_i(w)\triangleq\frac{1}{\lvert \mathcal{D}_i \lvert}\sum_{\xi_{i,j}\in \mathcal{D}_i}f(w,\xi_{i,j}),
\end{equation}
where $w$ is a parameter vector, which is typically synchronized with the server at the beginning of each round, and $f(\cdot)$ is a loss function. Formally, the objective is to solve:

\begin{equation} \label{eq2}
\underset{w}{arg\,min}\,F(w)\triangleq\sum_{i=1}^K \frac{\lvert \mathcal{D}_i \lvert}{\lvert \mathcal{D} \lvert} F_i(w),
\end{equation}
where $K$ represents the number of participating clients in each communication round. In round $r + 1$, client $c_i$ updates the parameter vector $w_i^r$ through mini-batch stochastic gradient descent (SGD) to minimize the objective function:

\begin{equation} \label{eq3}
    w_i^{r+1} = w^{r} - \eta_i \nabla F_i(w^{r}),
\end{equation}
where $w_i^{r+1}$ represents the model of client $c_i$ in round $r + 1$, $w^r$ denotes the global model in round $r$, and $\eta_i$ is the learning rate of client $c_i$. We use $g(w_i^r)$ to denote the gradient submitted by $i$-th client in round $r$. Upon receiving $K$ gradients, the server updates the global model parameters accordingly:

\begin{equation} \label{eq4}
    w^{r+1} = w^{r} - \eta \sum_{i=1}^Kp_i^rg(w_i^r).
\end{equation}
Here, $\eta$ is the learning rate of the server, $p_i^r$ denotes the weight of $i$-th client in round $r$, and $\sum_{i=1}^Kp_i^r=1$. 

\subsection{Enhancement of Gradient Stability (EGS)} \label{sec3.2}

FedHist incorporates a buffer at the server-side, comprising $\mathcal{G}$ and $\mathcal{L}$, which respectively cache global gradients and local gradients of past $h$ communication rounds from clients.

The goal of $\mathbf{EGS}$ is to select an appropriate collaboration gradient from $\mathcal{G}$ for each local gradient uploaded from a client, and then execute a weighted aggregation process. Quantitatively measuring the degree of similarity between two gradient vectors $g(w_i)$ and $g(w_j)$ can be achieved by computing their cosine similarity, as shown in Eq. (\ref{eq5}):

\begin{equation} \label{eq5}
  \mathbf{sim} \langle g(w_i),g(w_j) \rangle =\frac{{g(w_i)}\cdot{g(w_j)}}{\Vert{g(w_i)}\Vert\,\Vert{g(w_j)}\Vert}.
\end{equation}
We select the corresponding collaborative gradient for each local gradient based on the principle of minimizing similarity, as it implies a higher potential of sharing knowledge between two gradients and avoids overfitting \cite{hu2022fedcross}. This fine-grained cross-aggregation method mitigates gradient divergence and rectifies bias in local gradients. Thus the objective is to find the optimal collaborative gradient $g(w_{i,co}^r)$ in $\mathcal{G}$ for each $g(w_{i}^r)$:

\begin{equation} \label{eq6}
    \underset{g(w) \in \mathcal{G}}{arg\,min}\;\mathbf{sim}  \langle  g(w_i^r),g(w) \rangle. 
\end{equation}
The local gradient $g(w_i^r)$ is fused with the collaborative gradient $g(w_{i,co}^r)$ to enhance the diversity by Eq. (\ref{eq7}):

\begin{equation} \label{eq7}
    \tilde{g}(w_i^r) =g(w_i^r) + \alpha \, g(w_{i,co}^r),
\end{equation}
where $\tilde{g}(w_i^r)$ represents the local gradient after incorporating historical experiences, which will be utilized in the subsequent aggregation phase, and $\alpha \, \textgreater \, 0$ is a constant. Based on this, $\mathbf{EGS}$ alleviates client-drift by appropriately selecting a suitable collaborative historical global gradient for each local gradient, thereby enhancing its stability. 


\subsection{History-Aware Aggregation (HAA)} \label{sec3.3}

To promote fairness in FL, a key aspect is to identify potential contributors with more valuable data and allocate appropriate additional weights to them. With this goal in mind, $\mathbf{HAA}$ introduces a boosting factor to allocate higher weights to clients that exhibit outstanding performance in historically weighted aggregations. We combine the boosting factor with the exponential weighting method derived from \cite{chen2019communication}, using
$\lambda$ to control the weights of these two components:


\begin{equation} \label{eq8}
    \hat{p}_i^r =  \sum_{i=1}^K  \left(\left(\frac{e}{2}\right)^{-\tau(w_i^r)} + \lambda \cdot \overline{\boldsymbol{Util}}{(i,r)} \right), 
\end{equation}
where $\tau(w_i^r)$ and $\overline{\boldsymbol{Util}}{(i,r)}$ represent the staleness and utility value of the $i$-th client in round $r$, respectively. $\tau(w_i^r) \geq 1$ indicates the number of communication rounds have passed from the time the client $i$ received the last global model to the moment it starts uploading its new update and $\overline{\boldsymbol{Util}}{(i,r)}$ is obtained based on its historical behavior. And $\hat{p}_i^r$ needs to be normalized: ${p}_i^r =   { \hat{p}_i^r} / {\sum_{i=1}^K \hat{p}_i^r}.$ 

On the whole, building upon the concept of knowledge rumination, $\mathbf{HAA}$ employs the reminiscence of past data to generate unbiased gradients for prediction. It then compares the predicted values with the ground truth to calculate the client utility $\overline{\boldsymbol{Util}}$.  In fact, when $N$ is much higher than $K$, the lower bound approximation for the average staleness is close to $\frac{N}{2K}$ \cite{zhou2022towards}, which demonstrates that the local gradients uploaded by clients undergo a fixed number of rounds $ \left ( N /(2K)\right)$  of historical backtracking, resulting in obtaining corresponding relatively fresh updates.





\paragraph{Relatively fresh updates}
By configuring a historical buffer on the server-side, in round $r$, the server can trace back to historical gradients up to round $r-h$, thereby enabling the identification of relatively fresh updates. The so-called relatively fresh update refers to a gradient whose staleness is $h+1$ in round $r$, while its staleness is $1$ relative to round $r-h$. In other words, although it may be considered stale in round $r$, it is relatively fresh to round $r-h$. After the completion of each communication round, the server retrieves all local gradients from $\mathcal{L}$ within the buffer and stores all the gradients that satisfy Eq. (\ref{eq10}) into a pre-cleared set $\mathcal{S}$. 

\begin{equation} \label{eq10}
\begin{aligned}
    g_L \in \mathcal{L} \quad \text{s.t.} \quad  & \tau(g_L) = \text{index}(g_L, \mathcal{L}), \\
    \quad & 1 \leq \text{index}(g_L, \mathcal{L}) \leq h.
\end{aligned}
\end{equation}
Here, $\text{index}(g_L, \mathcal{L})$ denotes the index of the sublist containing $g_L$ within $\mathcal{L}$. Based on this, in round $r$, we are able to obtain all the relatively fresh updates to round $r-h$, and then we can obtain the predicted unbiased gradient.



\paragraph{Predicted unbiased gradients}
Since local gradients in $\mathcal{S}$ are relatively fresh to round $r-h$, their aggregated result has higher credibility to the global gradient in round $r-h$.
By performing aggregation on them, a predicted unbiased gradient $g_{pred}$ can be generated, representing the ideal global gradient obtained in the absence of staleness in round $r-h$.

\begin{equation} \label{eq11}
    g_{pred} \triangleq \frac{1}{\lvert \mathcal{S} \lvert}\sum_{g(w) \in \mathcal{S}} g(w).
\end{equation}
Note that the actual global gradient $g_{act}$ in round $r-h$, i.e., $g(w^{w-h})$, is still formed by aggregating those potentially stale local gradients. In fact, it is staleness that leads to the deviation between $g_{act}$ and $g_{pred}$. In order to assess the quality of the local gradients in round $r-h$, we can replay the previous knowledge to calculate their similarity with $g_{pred}$, followed by the corresponding client utility evaluation.


\paragraph{Client utility evaluation}
At the end of each communication round, the server evaluates the historical local gradients (referred to as $g_{his}^i$ hereafter for simplicity) participating in the aggregation phase in round $r-h$. $\mathbf{HAA}$ evaluates the utility of the corresponding clients using Eqs. (\ref{eq12}) - (\ref{eq13}).

\begin{equation} \label{eq12}
\begin{aligned}
\boldsymbol{Util}(i,r)= 
(\mathbf{sim}\langle g_{his}^i, g_{pred} \rangle - {sim}_{T}) \cdot p_{his}^i  \cdot \lvert \mathcal{S} \lvert,
\end{aligned}
\end{equation}
where ${sim}_{T}$ is a similarity threshold, and $p_{his}^i$ is determined based on the actual (not relative) staleness of $g_{his}^i$:

\begin{equation} \label{eq13}
\begin{aligned}
p_{his}^i = \begin{cases}
1-(e \, / \,2)^{-\tau(g_{his}^i)}  &  \\
\qquad \qquad \mathbf{if}  \quad \mathbf{sim}\langle g_{his}^i, g_{pred} \rangle \geq {sim}_{T}; \\
(e \, / \,2)^{-\tau(g_{his}^i)} &  \\
\qquad  \qquad \mathbf{if } \quad \mathbf{sim}\langle g_{his}^i, g_{pred} \rangle < {sim}_{T}.
\end{cases}
\end{aligned}
\end{equation}
According to Eqs. (\ref{eq12}) - (\ref{eq13}), the sign of $\boldsymbol{Util}(i,r)$ depends on whether the similarity exceeds the threshold (i.e., reward or penalty). If a gradient has a higher staleness during the actual aggregation process, it is expected to exhibit lower similarity with $g_{pred}$, thereby deserving relatively higher rewards or lower penalties, and vice versa. Moreover, the number of gradients in $\mathcal{S}$ reflects the predictive accuracy of $g_{pred}$, thus $\boldsymbol{Util}$ is positively correlated with $\lvert \mathcal{S} \vert$. To mitigate the influence of data noise, we employ an exponential moving average with a constant $\gamma$ to obtain the average utility $\overline{\boldsymbol{Util}}{(i,r)}$.

\begin{equation} \label{eq14}
    \overline{\boldsymbol{Util}}{(i,r)} = (1-\gamma) \cdot  \overline{\boldsymbol{Util}}{(i,r-1)} + \gamma \cdot \boldsymbol{Util}(i,r).
\end{equation}
By applying Eqs. (\ref{eq10}) - (\ref{eq14}) to Eq. (\ref{eq8}), $\mathbf{HAA}$ employs a multidimensional weighting approach based on the replay of historical gradients to explore stale yet more valuable gradient information, thereby promoting fairness.

\subsection{Intelligent  $\ell_2$-Norm Amplification (INA)} \label{sec3.4}
\underline{\textbf{Key observation}}. In K-async FL, the aggregation of local gradients often leads to a significant reduction in the $\ell_2$-norm of the aggregated result, which severely hinders the convergence process. Therefore, the directionality of the aggregated result holds more reference value, while its magnitude does not. Unfortunately, this observation is unexpectedly overlooked in many related methods \cite{chen2019communication}. To address this, we re-adjust the global gradients:


\begin{equation} \label{eq15}
g(w^r) = \frac{\sum_{i=1}^{K} g(w_i^r)}{\|\sum_{i=1}^{K} g(w_i^r)\|_2} \cdot \frac{1}{K} \sum_{i=1}^{K} \|g(w_i^r)\|_2.
\end{equation}
By modifying the $\ell_2$-norm of the aggregated result to match the average of all local gradients' before aggregation. This empowers $\mathbf{INA}$ to prevent reductions in the $\ell_2$-norm of the aggregated result. Generally, a larger $\ell_2$-norm of gradient implies the presence of more knowledge, and the $\ell_2$-norm of gradient learned from the previously trained samples tend to be relatively smaller \cite{shin2022fedbalancer}. Thus, in addition to smoothing the convergence process, $\mathbf{INA}$ also possesses the capability of perceiving the significance of updates.

\begin{table*}[htbp]
\centering
\caption{Test Accuracy (\%) when using FedHist and baselines on FMNIST, CIFAR-10, and CIFAR-100 Datasets. The best result in each column is shown in \textbf{bold} and the second-best is \underline{underlined}.}

\begin{tabular}{clccccccccc}
\hline
\multirow{2}{*}{Staleness} & \multirow{2}{*}{Methods} & \multicolumn{3}{c}{FMNIST} & \multicolumn{3}{c}{CIFAR10} & \multicolumn{3}{c}{CIFAR100} \\ \cline{3-11} 
                           &                           & $\beta=0.3$   &$\beta=1.0$   & IID    & $\beta=0.3$   & $\beta=1.0$   & IID     & $\beta=0.3$    & $\beta=1.0$   & IID     \\ \midrule
\multirow{8}{*}{$N/K=10$}    & FedAvg                    & 70.93  & 76.09  & 79.45 &  51.39 &  57.66  & 61.47  & 27.80  & 28.57  & 29.21  \\
                        & FedProx                   & 71.28  & 76.82 & 79.18& 52.41  &56.89 & 61.84 & 28.10  & 28.43  & \underline{30.75}  \\
                           & MOON                      & 71.72  & 76.39 & 79.35 &  52.08  &  57.50 &  61.39  &  27.90   & \underline{29.77}  & 30.34 \\
                           & DynSGD                    & 71.89  & 76.05  & 79.58 &  52.02  &57.90  &  62.11 & 27.75   & 29.06  & 30.33  \\
                           & REFL                      & 72.31  & 77.47  & 79.88 &  51.67  &57.72 &  61.76  &\underline{28.34}  & 28.77  & 29.03  \\
                           & TWAFL                     & 72.44 &74.67 & 79.48 & 51.43  & 57.71  & 61.91 & 27.27   & 29.26  & 29.73 \\
                           & WKAFL                     & \underline{73.45}  & \underline{77.92}  & \underline{80.04} &\underline{54.76}  & \underline{58.44} & \underline{62.26}  & 27.72   & 28.33  & 29.64  \\
                           & FedHist                   & \textbf{78.05}  & \textbf{81.34}  & \textbf{83.29} & \textbf{57.19}  & \textbf{59.78}  & \textbf{62.66} & \textbf{28.78}   & \textbf{29.89}  & \textbf{30.90}  \\ \hline
                           
                           
                           


\multirow{8}{*}{$N/K=100$}   & FedAvg                    & 70.24& 72.05  & 74.41 & 44.35  & 52.26  & 55.27  & 23.49   & 25.47  &24.81  \\
                           & FedProx                   & 71.01 & 72.40  & 74.11& 43.21 & 49.60  & 55.98 & 24.65   & 24.74  & 24.65  \\
                           & MOON                      & 71.79  & 74.94  & 75.85&41.59  & 50.72 & 53.57  & 23.10   & 22.69  & 22.77  \\
                           & DynSGD                    & 72.29  & 73.39 & 76.06 &51.11  & \underline{55.33}  & \underline{57.25} & 24.91   &\underline{26.14}  & 26.31  \\
                           & REFL                      & 70.98 & \underline{75.67}& \underline{77.28} &  \underline{51.46}  &  54.80 & 57.11& \underline{25.22}   & 25.76  & \underline{26.44}  \\
                           & TWAFL                     & \underline{72.73}  & 73.27  & 75.31& 47.16  & 53.00  & 55.87 & 23.53  &23.91  & 24.57  \\
                           & WKAFL                     & 70.56  & 75.37  & 74.97 & 45.60 & 51.42  & 55.01& 22.57   & 23.64  & 24.82  \\
                           & FedHist                   & \textbf{74.49} & \textbf{77.28} & \textbf{78.44} &\textbf{52.03} & \textbf{55.92} &\textbf{58.61} & \textbf{25.78}   & \textbf{26.31}  & \textbf{26.70} \\ \hline
\end{tabular}

\label{test_acc}
\end{table*}

\section{Experiments}
\subsection{Experimental Settings}
Our experiments are conducted on three well-known datasets, including CIFAR-10, CIFAR-100 \cite{krizhevsky2009learning}, and Fashion MNIST (FMNIST) \cite{xiao2017fashion}. To better simulate real-world scenarios of IID and Non-IID data, we employ the Dirichlet distribution $Dir_N(\beta)$ \cite{hsu2019measuring} to divide the dataset. Here, $\beta$ controls the degree of data heterogeneity, where a smaller $\beta$ signifies a more pronounced imbalance in the quantity and labels of samples. And we introduce different levels of staleness by setting $N/K$. We select several relevant algorithms for comparison: \textbf{FedAvg} \cite{mcmahan2017communication},  \textbf{DynSGD} \cite{jiang2017heterogeneity}, \textbf{TWAFL} \cite{chen2019communication}, \textbf{FedProx} \cite{li2020federated},  \textbf{MOON} \cite{li2021model}, \textbf{WKAFL} \cite{zhou2022towards} and \textbf{REFL} \cite{abdelmoniem2023refl}. All experiments are conducted using LeNet \cite{lecun1998gradient}.



\subsection{Performance Comparison}
\begin{figure}[t]
\centering
\includegraphics[width=1.0\linewidth]{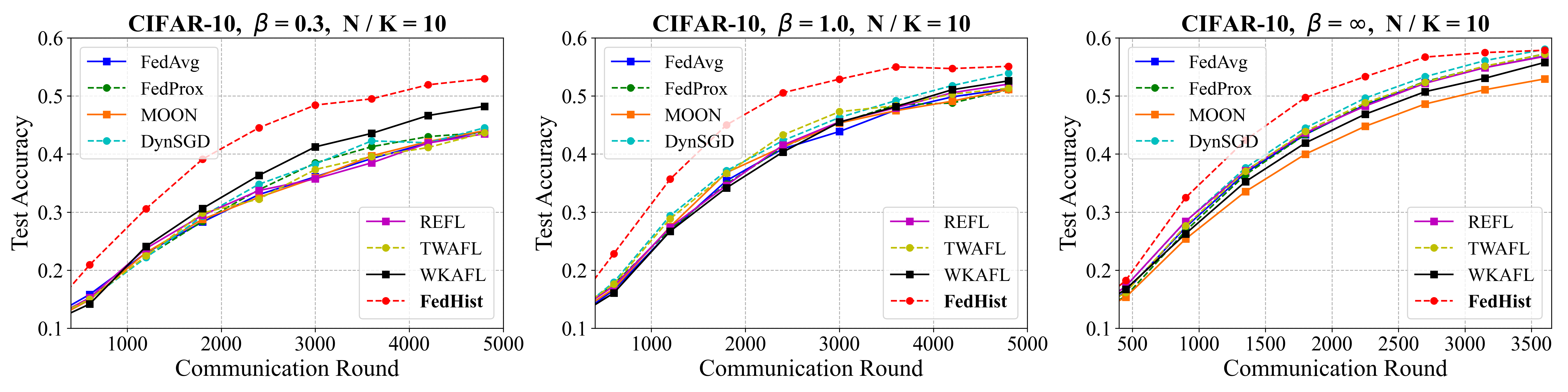}
\caption{Performance of FedHist and baselines during the training process on CIFAR-10 with $\beta$ = 0.3, 1.0 and $\infty$.}
\label{convergence_speed}
\end{figure}

\paragraph{Data heterogeneity}
Table \ref{test_acc} and Figure \ref{convergence_speed} illustrate that FedHist exhibits faster convergence rates compared to the baselines across three levels of heterogeneity, while also demonstrating the highest test accuracy, especially in highly heterogeneous environments. This is mainly attributed to the $\mathbf{EGS}$ within FedHist, which enhances the stability of local gradients and mitigates the impact of Non-IID data. Moreover, it is evident that the test accuracy of all methods improves with the increase of $\beta$, highlighting the detrimental impact of data heterogeneity on the training process.

\paragraph{Degrees of staleness}
Table \ref{test_acc} demonstrates the leading test accuracy of FedHist across various staleness levels. Although the test accuracy of all methods experiences a certain decline with the increase of $N/K$, the incorporation of the $\mathbf{HAA}$ in FedHist comprehensively considers the utility of client updates, thereby enabling its adaptability in highly stale environments and satisfactory robustness to staleness. Furthermore, \textbf{WKAFL}, \textbf{REFL}, and \textbf{TWAFL} achieve relatively commendable results owing to their concerted efforts in coping with staleness, while \textbf{FedAvg} performs the least favorably due to its lack of measures to address staleness.

\paragraph{Convergence speed}
Table \ref{speed} demonstrates that to achieve the target accuracy (40\%), FedHist achieves an acceleration ratio of approximately 1.8$\times$-2.0$\times$ compared to \textbf{FedAvg}, surpassing all baselines. The convergence trends of different methods displayed in Figure \ref{convergence_speed} intuitively highlight the advantages of FedHist in diverse data heterogeneity settings. 
Also, FedHist consistently gains a significant advantage in the early stages of training, and its convergence curve remains smoother. This is achieved by adaptively amplifying the $\ell_2$-norms of aggregated gradients through $\mathbf{INA}$. 


\begin{table}[!t]
    \caption{ Communication rounds in different methods to achieve target accuracy (40\%) on CIFAR-10. R\# represents the required number of communication rounds and S $\uparrow$ is the corresponding convergence speedup relative to FedAvg.}
    \centering
\begin{tabular}{lcccccc}
\hline

\multirow{2}{*}{Methods} & \multicolumn{2}{c}{$\beta$ = 0.3} & \multicolumn{2}{c}{$\beta$ = 1} & \multicolumn{2}{c}{$\beta$ = $\infty$} \\ \cmidrule{2-7} 
                         & R\#         & S $\uparrow$           &  R\#     & S $\uparrow$         &  R\#      & S $\uparrow$         \\ \hline
FedAvg                   & 3340        & 1×          & 2182         & 1×       & 1391          & 1×        \\
FedProx                  & 3876        & 0.9×        & 2370         & 0.9×     & 1562          & 0.9×      \\
MOON                     & 4324        & 0.8×        & 2088         & 1.0×     & 1444          & 1.0×      \\
DynSGD                   & 1904        & 1.8×        & 1573         & 1.4×     & 1176          & 1.2×      \\
REFL                     & 2177        & 1.5×        & 1289         & 1.7×    & 1013          & 1.4×      \\
TWAFL                    & 2695        & 1.2×        & 1492         & 1.5×     & 1244          & 1.1×      \\
WKAFL                    & 2954        & 1.1×        & 1701         & 1.3×     & 1299          & 1.1×      \\
FedHist                  & \textbf{1754}        & \textbf{1.9}×        & \textbf{1089}         & \textbf{2.0}×     & \textbf{792}           & \textbf{1.8}×      \\ \hline
\end{tabular}

\label{speed}
\end{table}

\begin{figure}[t]
\centering
\includegraphics[width=1.0\linewidth]{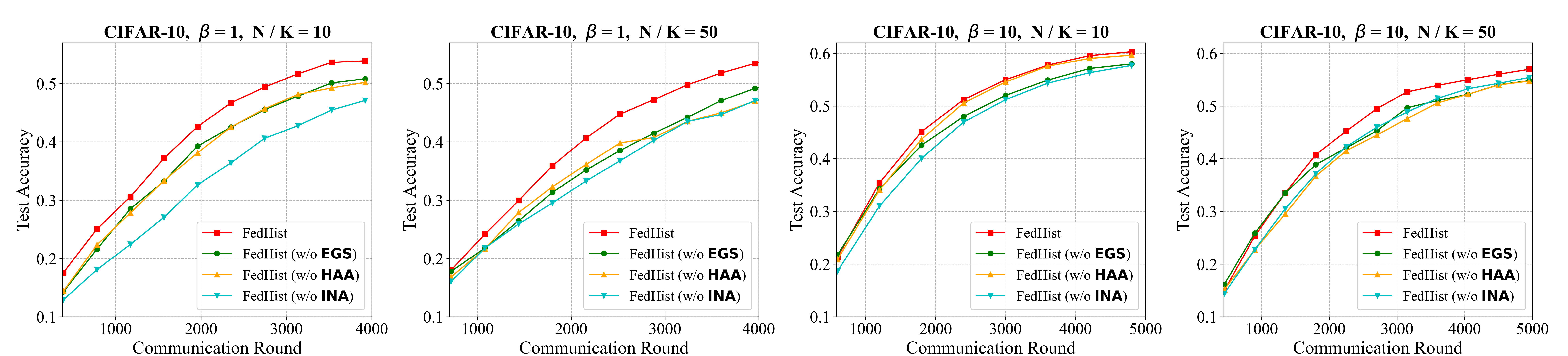}
\caption{Performance with different missing components on CIFAR-10.}
\label{ablation_study}
\end{figure}

\begin{figure}[t] 
\centering
\includegraphics[width=1\linewidth]{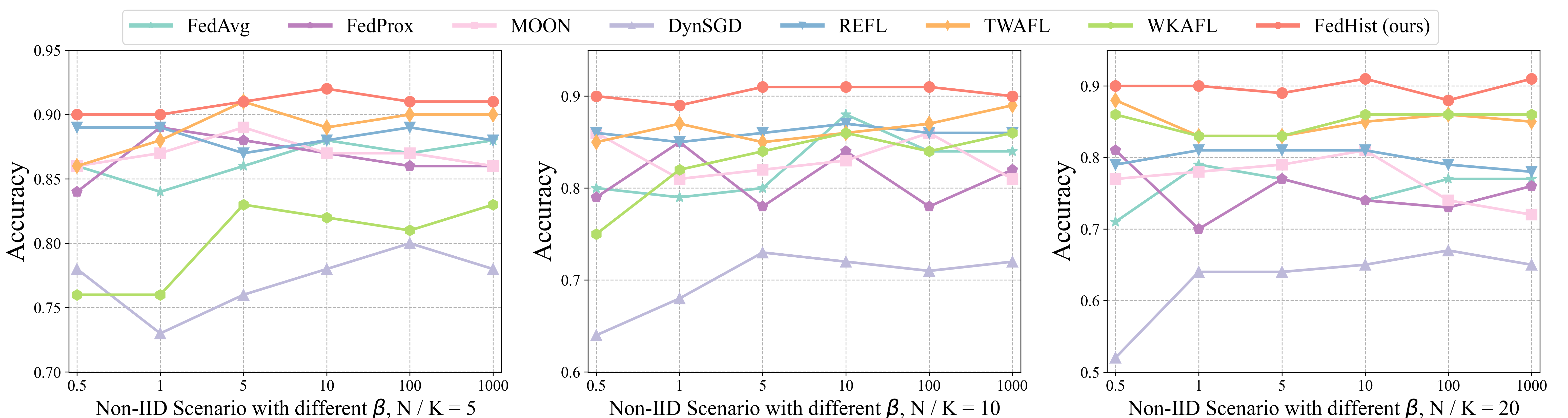}
\caption{Performance with specific labels occurring only on a few slow clients.}
\label{extreme}
\end{figure}

\subsection{Model Analysis}
\paragraph{Convergence analysis}
Table \ref{speed} compares the convergence rate of FedHist with the aforementioned baselines. The results demonstrate in both IID and Non-IID scenarios, FedHist achieves the fastest convergence rate to reach the target accuracy. 
Similarly, REFL also achieves results that are second only to FedHist by employing similar techniques. In contrast, DynSGD, TWAFL, and WKAFL exhibit relatively poorer convergence rates due to their focus on selecting fresh gradients while overlooking the value of stale gradients. The convergence trends of different methods displayed in Figure \ref{convergence_speed} intuitively highlight the advantages of FedHist. From these convergence plots, it can be observed that as $\beta$ increases, the convergence speed improves for all methods. 


\paragraph{Ablation studies} 
To further investigate the effectiveness of each component of FedHist, we conduct ablation studies, as shown in Figure \ref{ablation_study}. We systematically remove each of the three components and denote it as w/o (without) to indicate the absence of that particular component. The results demonstrate that regardless of the various data heterogeneities or different model staleness scenarios, FedHist consistently exhibits rapid convergence capability and stability. Indeed, the inferior stability observed in FedHist (w/o  $\mathbf{EGS}$) and FedHist (w/o $\mathbf{HAA}$) indicates the impact of Non-IID data and staleness leads to the deviation of local gradients, hinders the convergence of global model. The slower convergence speed observed in FedHist (w/o $\mathbf{INA}$) reflects the effectiveness of $\ell_2$-norm amplification strategy. 



\paragraph{Verification of fairness improvement} 
To validate the capability of FedHist in handling extremely heterogeneous scenarios, we conduct an experiment by setting the tenth class samples (for a 10-class classification task) to occur only in a small fraction of slow clients with an update frequency of 50\% compared to normal clients, and compare the experimental results of all the methods. As shown in Figure \ref{extreme}, FedHist demonstrates the highest test accuracy for these samples, which remains unaffected by changes in different degrees of data heterogeneity or staleness. This can be attributed to its consideration of aggregation weights beyond staleness. Through this weighting approach, FedHist can adapt to heterogeneous scenarios while ensuring that slower clients also contribute, thereby promoting fairness.

\section{Conclusion}
In this work, we propose a novel AFL method (FedHist) that replays the knowledge of historical gradients and mirrors the utility of clients. FedHist first enhances the stability of local gradients by fine-grained cross-aggregation. It assigns the aggregation weights in a multi-dimensional manner and enables fast and stable convergence by adjusting the $\ell_2$-norm of the aggregated gradient. We conduct extensive experiments on public datasets and validate its effectiveness.


\bibliographystyle{IEEEbib}
\bibliography{icme2025references}

\end{document}